\documentclass[10pt,twocolumn,letterpaper]{article}

\usepackage{cvpr}              

\usepackage[dvipsnames]{xcolor}

\definecolor{cvprblue}{rgb}{0.21,0.49,0.74}
\usepackage[pagebackref,breaklinks,colorlinks,citecolor=cvprblue]{hyperref}

\def\paperID{21} 
\def\confName{CVPR}
\def\confYear{2024}

\title{High-Resolution Detection of Earth Structural Heterogeneities from Seismic Amplitudes using Convolutional Neural Networks with Attention layers}

\author{
    Luiz Schirmer\textsuperscript{1},
    Guilherme Schardong\textsuperscript{2},
    Vinícius da Silva\textsuperscript{3},
    Rogério Santos\textsuperscript{4},
    Hélio Lopes\textsuperscript{3}\\
    \textsuperscript{1} Vizlab - X-Reality and GeoInformatics Lab, Unisinos\\
    \textsuperscript{2} University of Coimbra\\
    \textsuperscript{3} Pontifical Catholic University of Rio de Janeiro\\
    \textsuperscript{4} Federal Fluminense University \\
}

\begin{document}
\maketitle
\begin{abstract}
Earth structural heterogeneities have a remarkable role in the petroleum economy for both exploration and production projects. Automatic detection of detailed structural heterogeneities is challenging when considering modern machine learning techniques like deep neural networks. Typically, these techniques can be an excellent tool for assisted interpretation of such heterogeneities, but it heavily depends on the amount of data to be trained. 

We propose an efficient and cost-effective architecture for detecting seismic structural heterogeneities using Convolutional Neural Networks (CNNs) combined with Attention layers. The attention mechanism reduces costs and enhances accuracy, even in cases with relatively noisy data. Our model has half the parameters compared to the state-of-the-art, and it outperforms previous methods in terms of Intersection over Union (IoU) by 0.6\% and precision by 0.4\%. By leveraging synthetic data, we apply transfer learning to train and fine-tune the model, addressing the challenge of limited annotated data availability.
\end{abstract}    
\section{Introduction}
\label{sec:intro}

When geoscientists examine seismic data for hydrocarbon exploration in sedimentary basins, the features of structural heterogeneities, specifically fractures and faults, gain significant importance. These features possess distinctive characteristics that provide valuable insights into geological factors such as seal capacity, hydrocarbon migration, and control over relative fluid permeability. Obtaining fault and fracture information has been an important target for many scientific investigations to both pure research and economic viability studies\cite{di2018seismic}. Seismic methods traditionally used in the oil industry capture amplitudes that help generate interpretive models. These models are derived from frequency domains ranging from approximately 5Hz to 70Hz, which are economically distributed\cite{pochet2018seismic}. Detecting fractures using these frequencies is a challenging task that relies heavily on human expertise. This is because fractures are abundant and seismic resolution alone is insufficient to accurately define them with minimal uncertainty. The seismic industry has shown different levels of success for algorithms and artificial intelligence (AI) applied to either images or data, defining faults and fracture regions, seeking alternative workflows, and supporting hard human operational tasks \cite{pochet2018seismic,cunha2020seismic}. However, the intrinsic and hard limitations of algorithms' performance are related to how the description and conceptualization of faults and fractures should be modeled. The characterization of heterogeneities for oil prospecting is a key point when one wants to apply AI to fully establish and describe the geological concepts involved. Consequently, it is also important to better define specific computational methods applied to specific geologic targets. Here in this study, AI algorithms are designed for fault and fracture models and are exemplified along some combined geological scenarios involving salt tectonic and clastic depositional systems. Here, we propose an architecture for Fully CNNs combined with an attention module to enhance its capacity to analyze structural heterogeneities. Considering the learning task, we present a transfer learning (TL)\cite{zhuang2020comprehensive} strategy to train a network with synthetic data and tune a model to adapt it to real heterogeneity datasets. Our contribution is a simple yet effective architecture for seismic data analysis, which is not computationally intensive. In this approach, we start by working with a larger synthetic dataset initially and subsequently retrain the network while keeping certain layers frozen. Our results demonstrate that our model is highly competitive and effective, surpassing the previous state-of-the-art performance in terms of intersection over union (IoU) by 0.6\% and accuracy by 0.4\%.

\section{Related Work}

An important research goal in Computer Vision is modeling representations that focus on properties of data correlated only with the task at hand. Recent works have demonstrated that models represented by CNNs can be customized by the integration of learning mechanisms that specialize in capturing spatial correlations between features \cite{szegedy2015going, ioffe2015batch}. Other works focus on the expression of the spatial dependencies of a model \cite{bell2016inside, newell2016stacked} and incorporate spatial attention into the structure of the network \cite{jaderberg2015spatial}.


Attention for CNNs can be interpreted as a means of biasing the allocation of available computational resources towards the most informative components of a signal \cite{hu2018squeeze,olshausen1993neurobiological, itti1998model, itti2001computational,larochelle2010learning,mnih2014recurrent, bahdanau2014neural}.
Squeeze-and-Excitation Networks (SE-Nets) \cite{hu2018squeeze} is the first work to model cross-channel correlations. SE blocks synergize well with CNNs: they perform different roles for different depths of the network. In the first layers, the resulting excitation is informative but class-agnostic. Deeper blocks in the network become more specialized and produce a highly class-specific result being accumulated through the entire network \cite{hu2018squeeze}. The SE-Nets are also lightweight, imposing a small increase in model and computation complexity.

More recently, Bello \textit{et al.}~\cite{bello2019attention} proposed 2D self-attention as a replacement for convolutions for image classification tasks, since self-attention captures global behavior more appropriately than convolutional layers, which are inherently local. However, his global form attends to all spatial locations of an input, limiting its usage to small inputs which typically require significant downsampling of the original image. In our experiments, we apply those techniques to small seismic image patches and do not represents a major increase in the computation complexity of our model.

Considering applications of neural networks for seismic data analysis, Pochet \textit{et al.} \cite{pochet2018seismic} explore the use of Convolutional Neural Networks (CNNs) for seismic fault detection. One of the challenges associated with CNNs is the requirement for a large volume of interpreted data, which is particularly difficult to fulfill in the seismic domain. To address this challenge, the authors created a synthetic dataset focused on simple fault geometries. The input to their CNN model comprises only seismic amplitude data, eliminating the need for computing additional seismic attributes. This approach involves employing a patch classification strategy across the seismic images. This strategy entails a straightforward post-processing step to extract the precise location of faults. In contrast to our approach, we do not apply a post-processing step to the seismic data.

Cunha \textit{et al.} \cite{cunha2020seismic} suggest utilizing Transfer Learning techniques to leverage an existing classifier and apply it to different seismic data with minimal effort. Their foundational model is a Convolutional Neural Network (CNN) trained and optimized using synthetic seismic data. They showcase the outcomes of Transfer Learning specifically on the Netherlands offshore F3 block in the North Sea \cite{F3}. Our proposal shares similarities with the Transfer Learning approach; however, our model is not a classifier but a fully convolutional neural network. Additionally, aiming to propose approaches to interpreting seismic faults in seismic data, Palo \textit{et al.} \cite{palo2023fault} propose utilizing a graph neural network and Wang \textit{et al.} \cite{wang2023transformer} propose a Transformer-assisted dual U-Net. As we will show in our experiments, our models surpass the previous works in terms of accuracy and Intersection over Union. In summary, our study highlights the potential of CNNs for automating seismic fault detection, offering promising results both on synthetic datasets and in real-world applications.

\section{Geoscientific Background}

Our AI workflows was applied to some geoscientific scenarios and illustrated for F3 Block in Dutch Sector in North Sea Netherlands, whose 3D seismic data is kindly made available by DGB Earth Sciences\cite{F3}. A Post Stack Time Migration (PSTM) with a very good quality presents seismic data of Cenozoic layers greatly affected by salt tectonics and by depositional characteristics of clastic sedimentation, where the Eridanos delta stands out economically. Here, concepts are described and exemplified along three Inlines of such PSTM, illustrating some of main geological scenarios in the F3 Block.


The study area is located in a sedimentary basin in the south of North Sea, tectonically described by two regional phases: a rifting during the Cretaceous and a sag post-rift in the Cenozoic. Here, the unconformity between the Cretaceous and Cenozoic represents the oldest available surface for seismic reflections.
We perform tests with sections of available PSTM, following the work of Schroot and Sch\"{u}ttenhelm \cite{schroot2003expressions}. They described the seismic effects of gas occurrences and their origins in the studied area. Two of their cases are here analyzed, over which our algorithm is applied, showing the benefits for structural heterogeneities understanding. The first case refers to bright spots caused by the seismic effect of gas presence in anticlines that reach diameters with dozens of kilometers which could become economically interesting. However, it would not be a reality for all seismic responses of these gases in the study area. A question then arises: how to measure these accumulations and bring better evidence of the actual and economic presence of gas accumulation?

\begin{figure}[h]
\centering
\includegraphics[width=1.0\columnwidth]{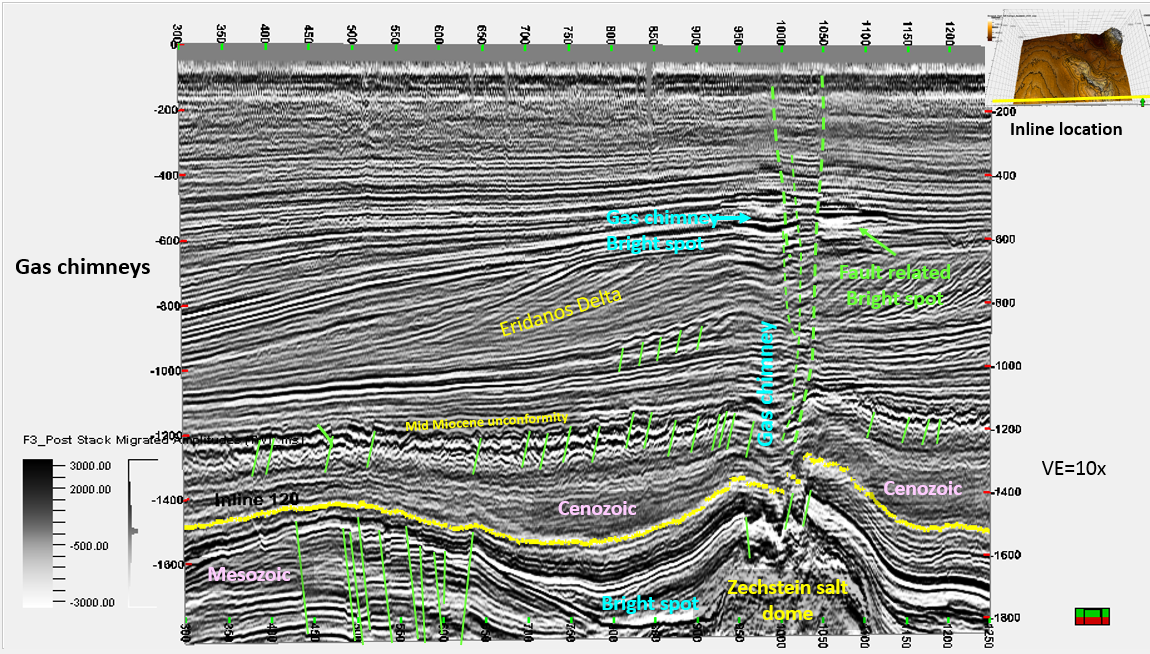}
\caption{Seismic Inline 120 within TWT (ms) showing the  base of Cenozoic reflection (yellow line) in F3 Block. A gas chimney adjacent to a fault, and immediately overlying an apparently leaking bright spot at reservoir levels.  A seismic pull-down effect can be seen underneath the chimney. Green segments can be modeled as faults}
\label{fig_1}
\end{figure}

A common type of seismic anomaly present in F3 Block is related to gas chimneys, caused by different mechanisms that are manifested through disturbances in the seismic signal related to the upward movement of the fluids residing in the pores, reducing elastic properties, generating wave scattering and defocusing the reflected energy. Figure \ref{fig_1} illustrates Inline 120 in TWT(ms) and shows with yellow line the deepest available horizon of the post-rift sag phase for this work representing the Mid-Miocene unconformity. The seismic response of a gas chimney is located above the salt dome in the Zechstein formation, which tectonically builds fracture and fault zones through which the gas can flow and accumulate within possible adjacent reservoirs. Schroot and Sch\"{u}ttenhelm \cite{schroot2003expressions} mention that the continuity of the reflections along with the chimney, in Cenozoic sediments, may show a slow and moderate gas migration causing pulldown seismic effects. For all these sediments, the effects of salt tectonics are evident, and it is here shown how these effects can be automatically highlighted by our AI method, seismically enhancing that gas migration can occur through fractures in the internal space defined in the chimney itself.

\begin{figure}[h]
\centering
\includegraphics[width=1.0\columnwidth]{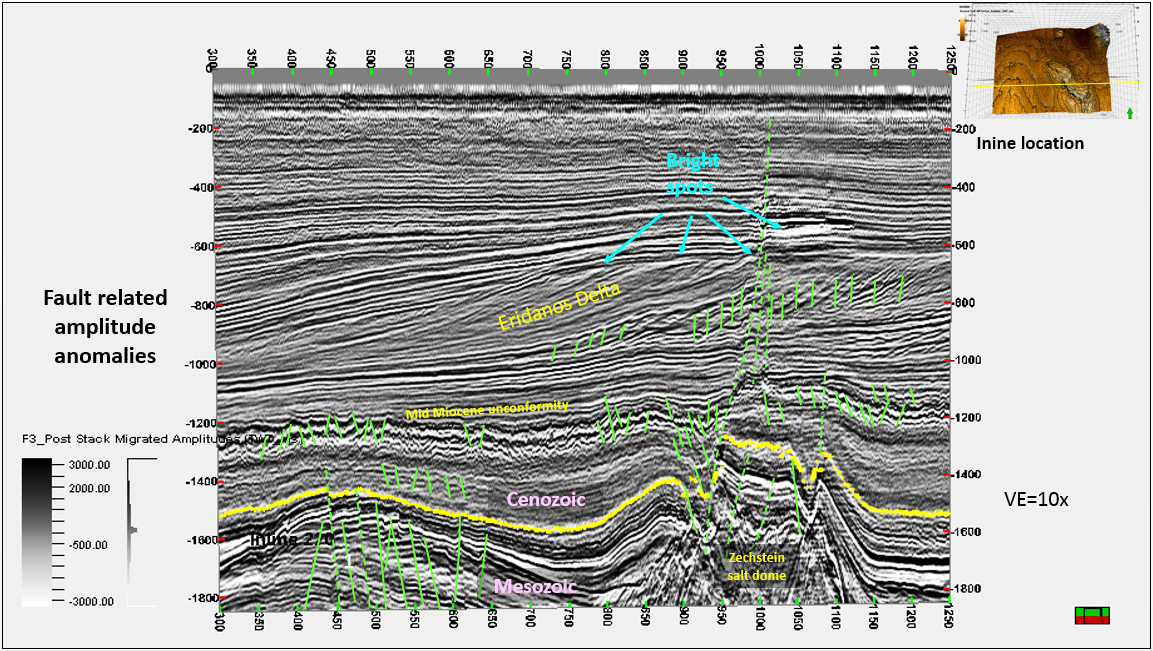}
\caption{Seismic Inline 220 within TWT (ms) showing the base of Cenozoic reflection (yellow line). Leakage suggestions are indicated as fault systems in the northern part of F3 block. Faults provide gas migration paths and control bright spot indicators. All green segments can be modeled as faults.}
\label{fig_2}
\end{figure}

A second case here described for amplitude anomalies is related to structural heterogeneities suggesting gas migration pathways in Inline 220 of F3 Block, illustrated in Figure \ref{fig_2}. A single and stronger bright spot is comprehensively related to gas leaks through extension structural heterogeneities. Some of them extend over long trajectories in the stratigraphic section, reaching the ocean floor and showing small patches of high seismic amplitudes throughout the stratigraphy. Important to observe that many heterogeneities modeled internally to the Eridanos delta are not so easy to be visualized with amplitude data. With our results, it is possible to suggest that gas migration paths can reach the seal unconformity at the top of the delta and then accumulate in small patches, like unassuming bright spots.

\begin{figure}[h]
\centering
\includegraphics[width=1.0\columnwidth]{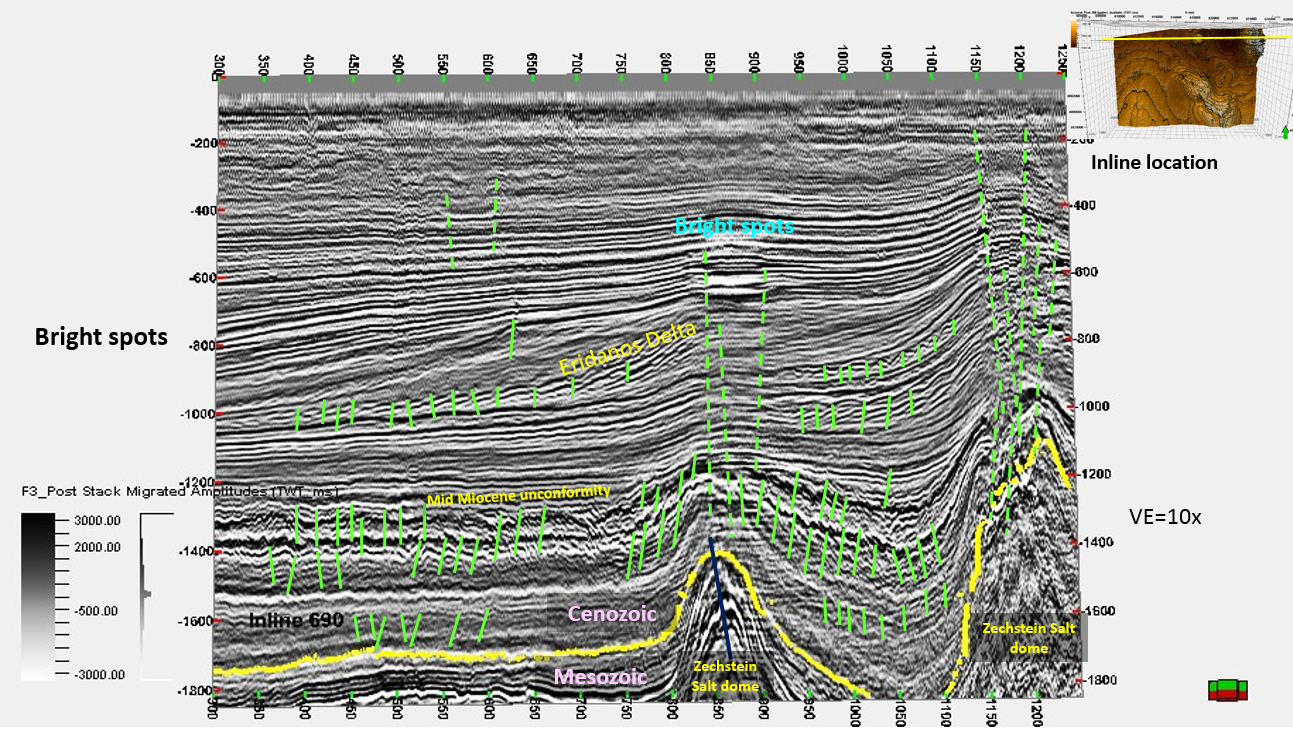}
\caption{Seismic Inline 690 within TWT (ms) showing the base of Cenozoic reflection (yellow line) in F3 Block. A shallow bright spot over a flat spot corresponds to the gas-water contact in the Upper Pliocene sediments. This seismic expression is indicative of effective structural gas trapping. Green segments could be modeled as faults} \label{fig_4}
\end{figure}

Figure \ref{fig_4} illustrates the second case, related to PSTM in Inline 690. A shallow bright spot is shown over a flat spot, which would occur close to a structural trap. Here, structural heterogeneities describe paths that the gas would take until reaching the trap and the role that fractures and faults have in gas migration and retention. About a hundred segments of possible faults are indicated with green color. Some of them would lead gas to traps, and others could break seals. Despite be possible to model those segments as faults, all heterogeneities here shown are not necessarily related to the structural origin and may have either sedimentary or diagenetic origins. Mapping hundreds of structural heterogeneities, in all lines of interest, is an exhaustive, time-consuming, and sometimes economically uninteresting task along E$\&$P chain, once for the exploration phase we would be generically interested in major elements of petroleum systems and for reservoir characterization the focus is basically more related to small connectivity, segmentation of reservoirs and permeability barriers.

\section{Methods and Network Architecture}

\subsection{Convolutional model and synthetic dataset}

Here we aim to find patterns in amplitude patches. We generate as output "confidence maps" indicating if the region of the analyzed image has any type of heterogeneity, such as faults or fractures. In addition, the theory related to CNNs approaches to the wavelet transforms, which is widely used in seismic interpretation. In seismic analysis, spectral decomposition and continuous wavelet transform were used in data processing, interpretation, and for evaluation of hydrocarbon reservoirs. Convolutional neural networks (CNNs) are effective in learning filters to detect patterns in seismic data. CNNs can learn hierarchical representations of data through convolutional layers and these layers capture local patterns and features, which can be seen as a form of localized linearization. 
While both CNNs and wavelet transforms aim to capture patterns at different scales, their underlying mechanisms differ significantly. Wavelet transforms are based on mathematical functions that analyze signals in both the frequency and time domains, providing insights into localized changes in the signal. In contrast, CNNs learn filters through backpropagation and gradient descent, adapting them to capture relevant features from the data during training \cite{mallat2016understanding}.


To obtain our initial training dataset, we adopt a strategy similar as proposed by Pochet \textit{et al.}\cite{pochet2018seismic} with the IPF code from Hale \cite{Hale2014}. As said before one of the major problems in training a CNN structural heterogeneity detection is the lack of annotated data. This machine-learning technique depends on a large amount of annotated information to generalize the solution. To solve this problem, we use the toolkit published by Dave Hale \cite{Hale2014} for seismic image processing for geologic faults to generate synthetic data. We initiated the process by creating a reflectivity model spanning the section, and then applied simple image transformations to simulate sequential rock deformations such as shearing, folding, and faulting over time. This process was followed by a convolution with a Ricker wavelet and the addition of random noise \cite{pochet2018seismic,cunha2020seismic}. 

\begin{figure}[!t]
\subfloat[Synthetic Seismic image]{\includegraphics[width=1.0\columnwidth]{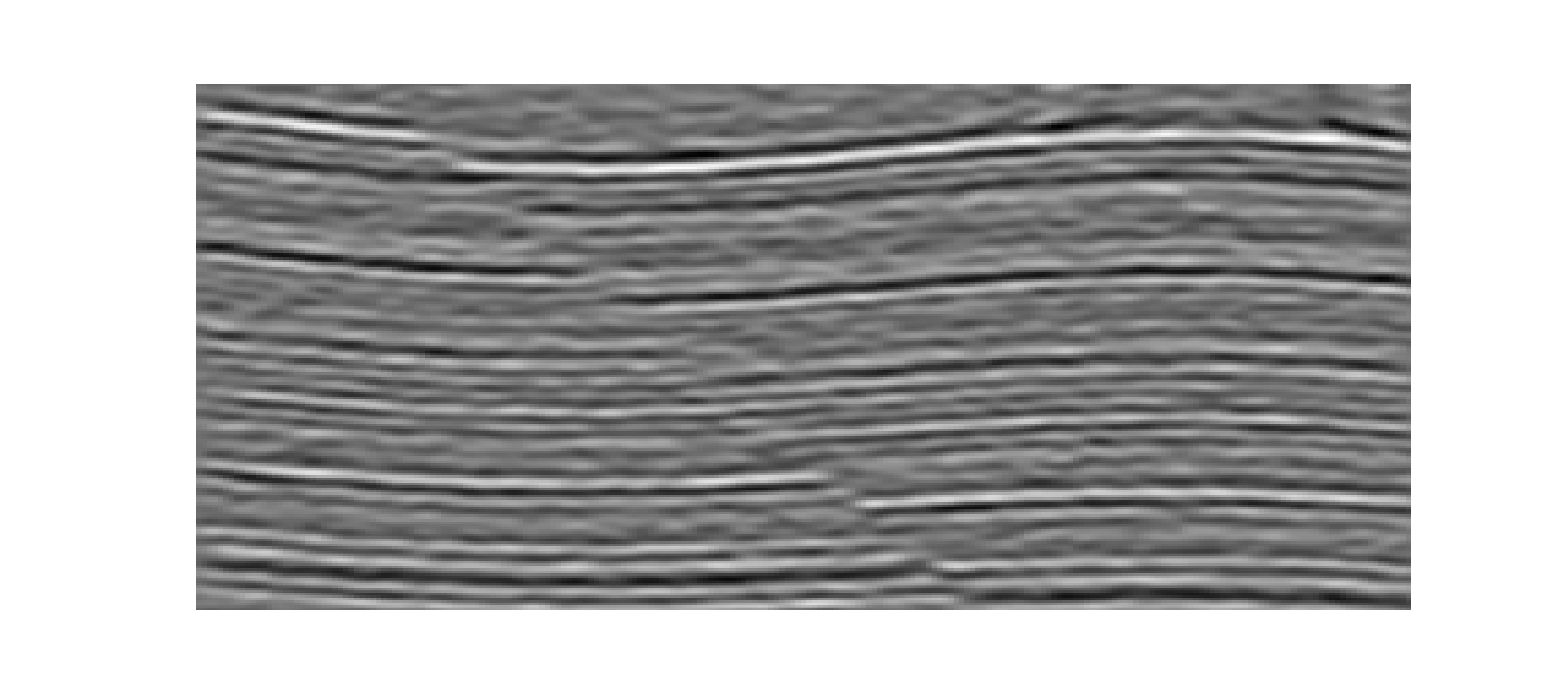}}
\label{Synthetic image}

\subfloat[Annotation]{\includegraphics[width=1.0\columnwidth]{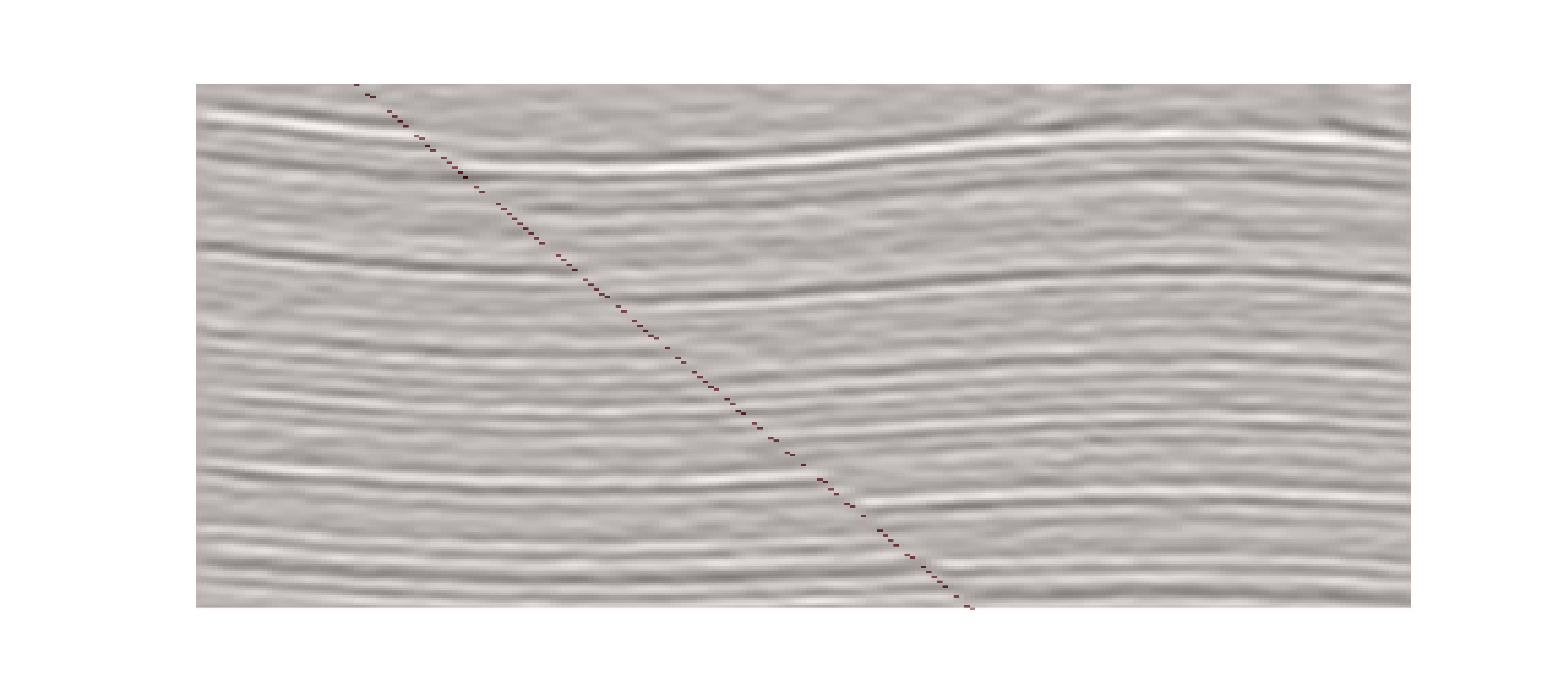}
\label{Annotation}}
\caption{The image at the top is a synthetic image generated using the IPF code developed by Hale\cite{Hale2014}. Below it, we have an image with a dashed line representing fault data annotations. During training, these annotations are converted into a binary mask.
}
\label{fig_anotation}
\end{figure}

We produce approximately 4000 images which we subdivide into patches of size $44\times44$ pixels, with amplitude values normalized from -1 to 1, where for each patch we have a confidence map representing the region with Heterogeneities and Non-heterogeneities, i.e, regions with seismic faults and without~\cite{cunha2020seismic,pochet2018seismic}. This technique was used as a basis for our initial training process. As the result, we have a synthetic dataset of seismic images and a set of grounth-truth binary masks indicating the presence of faults. The IPF code only generates fault patterns. However, we train our network to detect faults, and after, in a transfer learning process, we adapt it to detect other heterogeneities. As we will present later, we use a small annotated dataset with this information and we try to generalize the learned information to other domains. The dataset was split between 80\% of the data for training and 20\% for testing. The figure \ref{fig_anotation} presents an example of the synthetic data generated and its annotation.

\begin{figure}[h]
\centering
\includegraphics[width=1\columnwidth]{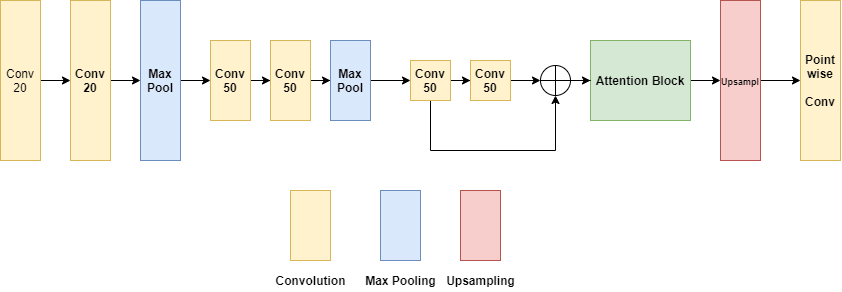}
\caption{Architecture of our neural network. Our model is based on a fully convolutional architecture where similarly to DenseNets we have a residual operation before the attention block. We generate two versions of this architecture, one using SE-Nets and the other using a Self-attention block.}
\label{fig_CNN}
\end{figure}

Our network is almost entirely composed of convolutions and max pooling operations. Two convolution layers, with a kernel size of $3 \times 3$ and 20 channels compose the first stage. The second and third stages have 2 convolutional blocks with a kernel size of $3 \times 3$ and 50 channels. These convolutional stages are followed by an attention block. In our experiments, we test both spatial and channel-wise attention base on SE-Nets\cite{hu2018squeeze} blocks and a self-attention model\cite{bello2019attention}. As input for the attention module, we concatenate the output of the previous two convolution layers, as in a residual skip connection. After the attention block, there is an upsampling operation set and a final pointwise convolution is performed producing a two channel output. This upsampling operation is performed by a simple deconvolution layer with a $3\times3$ kernel size.  The objective of our network is to assign a probability of a class to each pixel in the input image. To achieve this, the output image predicts the probability of two classes for the pixels: heterogeneities and non-heterogeneities. Figure \ref{fig_CNN} presents the architecture of our neural network. We also use GeLU as our activation functions and an cross-entropy loss between the estimated predictions and ground truth "confidence maps". We adopt the Adam optimizer \cite{kingma2014adam} with a learning rate of 0.001 and we trained this model by 200 epochs (considering each of attention versions).

\subsection{Attention block}
In this section, we present the main aspects of the attention block used. We use two different models for this block and here we will present the main aspects of each one.

\subsubsection{Squeeze-and-excitation networks}
The original SE-Net proposed a ``Squeeze-and-Excitation'' block~(hence, SE-Net) to highlight the channel-wise feature maps by applying weights to each channel~\cite{hu2018squeeze}. Following the SE-Net architecture in the squeeze operation, a global average pooling is defined by Equation~\ref{eq.1}\cite{hu2018squeeze,su2019multi} generating the weights $z$:

\begin{equation}
    z_{c} = \frac{1}{H \times W} \sum_{i=0}^{H} \sum_{j=0}^{W} X_{c}(i,j), 
\label{eq.1}
\end{equation}

\noindent where $X_{c}$ corresponds to each channel of our input. After, in the excitation step, we fully capture channel-wise dependencies, and a sigmoid function~($S$) is applied. Consider an input with $ch$ channels. This mechanism is composed of two fully connected layers (FLC) where the first has  $\frac{ch}{r}$ neurons and a GeLU activation and the second $ch$ neurons. Finally, the second FLC is followed by the sigmoid activation. The hyperparameter ratio $r$ defines the capacity and computational cost of the SE blocks in the network and can be evaluated empirically~\cite{hu2018squeeze}. In the sequence, to generating the output, a Hadamard product between the result of the gating mechanism and data input is performed. Formally, the attention block is defined by Equation \ref{eq.2}:

\begin{equation}
\mathcal{X^{'}} = \sigma(W_2(\beta (W_1 z))) \circ \mathcal{X},
\label{eq.2}
\end{equation}

\noindent where $W_1 \in \mathbb{R}^{\frac{ch}{r} \times ch}$ and $W_2 \in \mathbb{R}^{ch \times \frac{ch}{r}}$ represents two fully connected layers, $\beta$ is a GeLU activation and $\sigma$ is a Sigmoid activation function. Also, the element-wise multiplication is performed over the input $\mathcal{X}$. Similarly, to Su et al.~\cite{su2019multi}, after, with the channel-wise attention, we also perform spatial attention over the output $\mathcal{X^{'}}$. 




\subsubsection{Self-attention}
\label{sec:attn-conv}

To create our second attention module, we use a two-dimensional relative self-attention mechanism based on the Transformer architecture. This approach was proposed by Bello et al.~\cite{bello2019attention} where the use of multi-head attention~(MHA) allows the model to attend jointly to both spatial and feature subspaces, also producing additional feature maps. Let $H$, $W$ and $F_{in}$ as the height, width and number of input filters of an activation map; $Nh$, $dv$ and $dk$ as the number of heads, the depth of values, queries and keys in a multihead-attention~(MHA), and $d_v^h$ and $d_k^h$ the depth of values and queries/keys per attention head. With a tensor $\mathcal{T}$ with shape $(H,W,F_{in})$, they produce a flattened form with dimensions ($H \times W$, $F_{in}$) defined by the matrix $X$ and perform MHA. In this process they also have $W_q, W_q \in \mathbb{R}^{F_in \times d_h^k}$ and $W_v \in \mathbb{R}^{F_in \times d_h^v}$ that are learned linear transformations that maps the matrix $X$ to queries $Q=XW_q$, keys $K=XW_k$ and values $V=XW_v$.  Relative height information and relative width information were added and the relation between a pixel $i$ and a pixel $j$ is defined by ~Equation~\ref{eq.4}~\cite{bello2019attention}:

\begin{equation}
    \text{logit}_{i,j} = \frac{q_{i}^T}{\sqrt {d_k^h}} (k_j + r_{j_x - i_x}^W + r_{j_y - i_y}^H)
    \label{eq.4}
\end{equation}

\noindent where $q_i$ is a query vector for a pixel i in $Q$ and $k_j$ is a key vector for the pixel $j$ in $K$. Also $r_{j_x - i_x}^W$ and $r_{j_y - i_y}^H$ are learned embeddings for relative width $j_x - i_x$ and relative height $j_y - i_y$. The output of the self-attention for a single head $h$ is defined by Equation~\ref{eq.5}~\cite{bello2019attention,vaswani2017attention}:

\begin{equation}
    O_h = \kappa \left(\frac{QK^T + S_W^{rel} + S_H^{rel}}{\sqrt d_k^h}\right)V 
    \label{eq.5}
\end{equation}

\noindent where $S_W^{rel}$ and $S_H^{rel}$ are matrices of relative position logits along height and width dimensions that satisfy $S_H^{rel}[i,j] = q_i^T r_{j_y - i_y}^H$ and $S_W^{rel} = q_i^T r_{j_x - i_x}^W$~\cite{bello2019attention} and $\kappa$ is Softmax activation function.

The outputs of all heads are then concatenated and projected to generate the Multi-head attention and the final attention augmented as in equations~\ref{eq.6} and \ref{eq.7}. 

\begin{equation}
    \text{MHA} = \text{concat}[O_1, \dots , O_{Nh}]W^O
    \label{eq.6}
\end{equation}

\begin{equation}
    \text{AAC}(X) = \text{conv}(X)\oplus\text{MHA} 
    \label{eq.7}
\end{equation}

\noindent where $W^O$ is a learned linear transformation. In total considering the convolution operations, Multihead attention uses a $1\times1$ convolution with $F_{in}$ input filters and $2d_k + d_v = F_{out}(2k+v)$ output filters to compute queries, keys and values and an additional $1 \times 1$ convolution with $d_v=F_{out}v$ input and output filters to obtain the values and combine different heads outputs.

\subsection{Transfer Learning Approach}

In the transfer learning process, similarly to Cunha et al. \cite{cunha2020seismic} and Pochet et al.\cite{pochet2018seismic}, we need to preprocess the real data to be used. We use a dataset with annotated information for seismic faults and heterogeneities from the F3 North Sea. Note that at the first stage considering the limitations of the IPF code \cite{Hale2014}, with this real data we intend to improve the capacity of this neural network to consider also heterogeneities and not only seismic faults. The base model was trained using synthetic patches that do not match the frequency of real seismic data\cite{cunha2020seismic}. Real seismic signals typically range between 20 and 100 Hz, but the synthetic dataset lacked patches with frequencies higher than 20 Hz. This caused differences in amplitude patterns between synthetic and real patches when using the same patch size for both datasets. For training, we select 5 seismic inlines from the F3 North Sea manually interpreted and annotaded by a geophysicist, more specific the inlines 100, 150, 250, 300 and 600. We extract patches from annotated seismic images considering a size $20\times20$ and re-scale them to the size of $44\times44$, which naturally lowered their seismic signal frequency. This adjustment helped align the characteristics of real and synthetic patches, improving the performance of our model across both types of seismic datasets. The annotated data corresponds to binary masks of each inline, where regions containing heterogeneities were manually annotated. This ground-truth data was also divided into patches of $44\times44$.

After using the transfer learning approach, we continuously use this data to train our neural network. We previously extracted information about heterogeneities from the synthetic dataset and adjusted the layers of the neural network for real data. We freeze the weights for the first two convolutional layers and then we retrain our network by 30 epochs and we use a learning rate of 0.001 . This mitigates the problem of not having enough data to train complex neural networks. In practice, we adjust the weights of the refinement stage, including the second and third convolution layers, the attention block, and the final convolution layer. We adopt and tested this strategy for the two versions of our neural network, one using the SE-Net as our Attention Block and the other one using the Self-Attention. For testing, we select three inlines from the F3 dataset, specifically inlines 120, 220, and 690. To ensure a fair comparison, we also evaluate our model using different data, specifically a subvolume of the New Zealand Great South Basin dataset (GSB)\cite{di2017seismic} spanning crosslines 2568 to 3568. We evaluate the results against the state-of-the-art methods and compared the results obtained with the interpretation made by a geophysicist.
\section{Experiments}
 Table~\ref{tab:table_1} shows the results of our method compared to the state-of-the-art methods found in the literature. As we can our best model has a IoU of $91.2\%$ and a precision of $95.7\%$. Our model exhibits higher accuracy compared to other approaches, enabling it to detect more detailed heterogeneities in practice. As shown in Table \ref{tab:table_1}, both our models surpass previous works. Additionally, our models are compact in terms of storage and memory usage, and efficient enough to be trained on a low-cost device in less than 15 minutes. In our experiment, we utilized an Nvidia RTX 2060 GPU with 6GB of memory. The computing power required to deploy our model reached a maximum of 791.356 KiloFLOPS, and the model consists of 92827 trainable parameters. Considering our architecture, in comparison with approaches utilizing U-net based models and Graph convolutions, our model has fewer than half the parameters.

\begin{table*}[!t]

\centering

\caption{Quantitative evaluation of our model and previous works using the F3 Dataset. Parameters were tuned manually.}
\label{tab:table_1}
\begin{tabular}{lrrrr}
\hline
Model                      & \multicolumn{1}{r}{IoU} & \multicolumn{1}{r}{Recall} & \multicolumn{1}{r}{Precision} & \multicolumn{1}{r}{F1}\\ \hline
SE-NET (Ours)                                                   & 0.712          & 0.968                         & 0.662 & 0.78\\
Self-attention (Ours)                                                  & \textbf{0.912}          & 0.984                         & \textbf{0.957} & \textbf{0.97}\\

SVM \cite{cunha2020seismic}                           & 0.572                   & 0.966                         & 0.60    &   0.78                     \\
MLP \cite{cunha2020seismic}                                    & 0.609                   & \textbf{0.997}                & 0.683   & 0.81                        \\
Fine Tuning \cite{pochet2018seismic}                             & 0.554                   & 0.949                         & 0.564   & 0.70                        \\ 

GCN \cite{palo2023fault}                              & 0.906                   & 0.48                         & 0.85    & 0.61                       \\

Transformer dual U-Net \cite{wang2023transformer}                            & 0.90                   & 0.83                         & 0.954   & 0.88                        \\ \hline

\end{tabular}

\end{table*}

\begin{figure}[!t]
\centering
\subfloat[Interpreted annotations provided by Haibin Di et al\cite{di2017seismic}]{\includegraphics[width=3in]{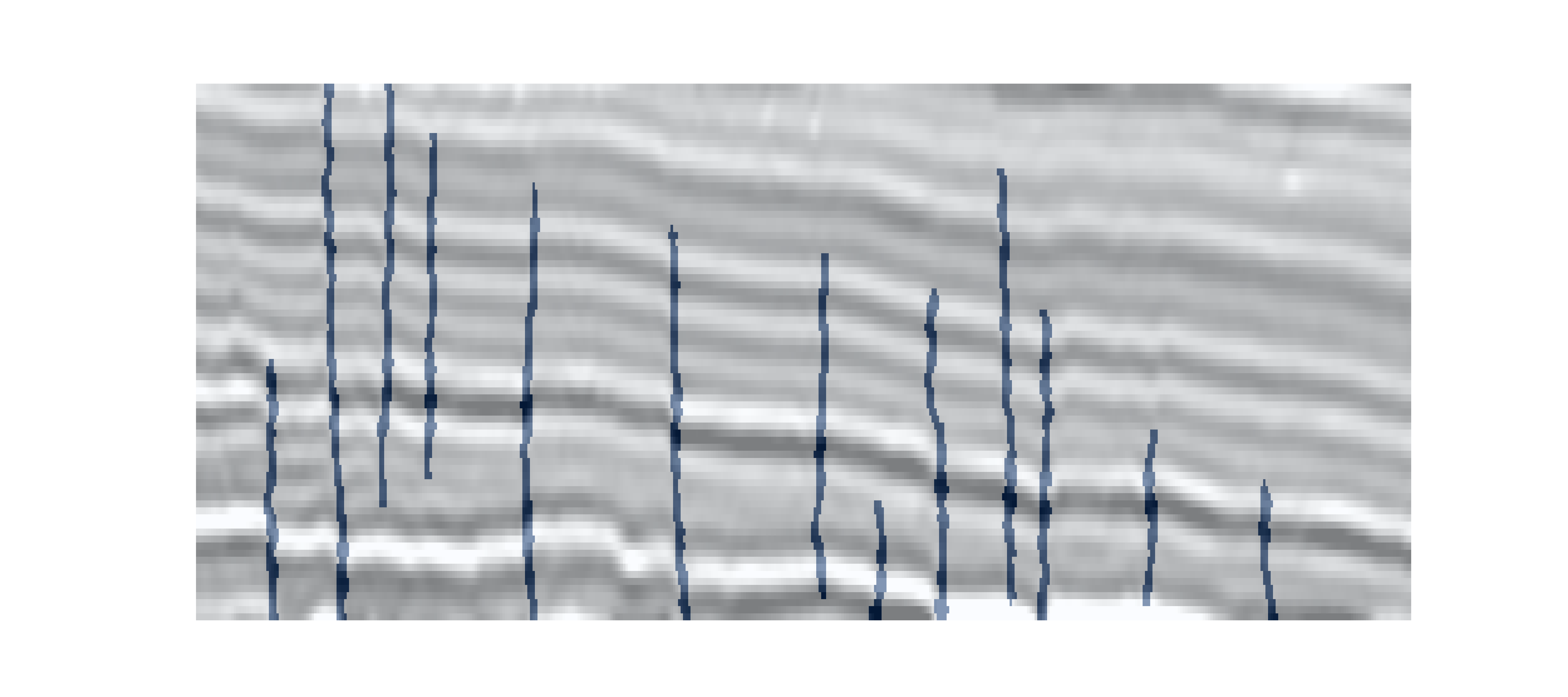}}
\label{fig_first_case_ts}

\subfloat[Our Model results]{\includegraphics[width=3in]{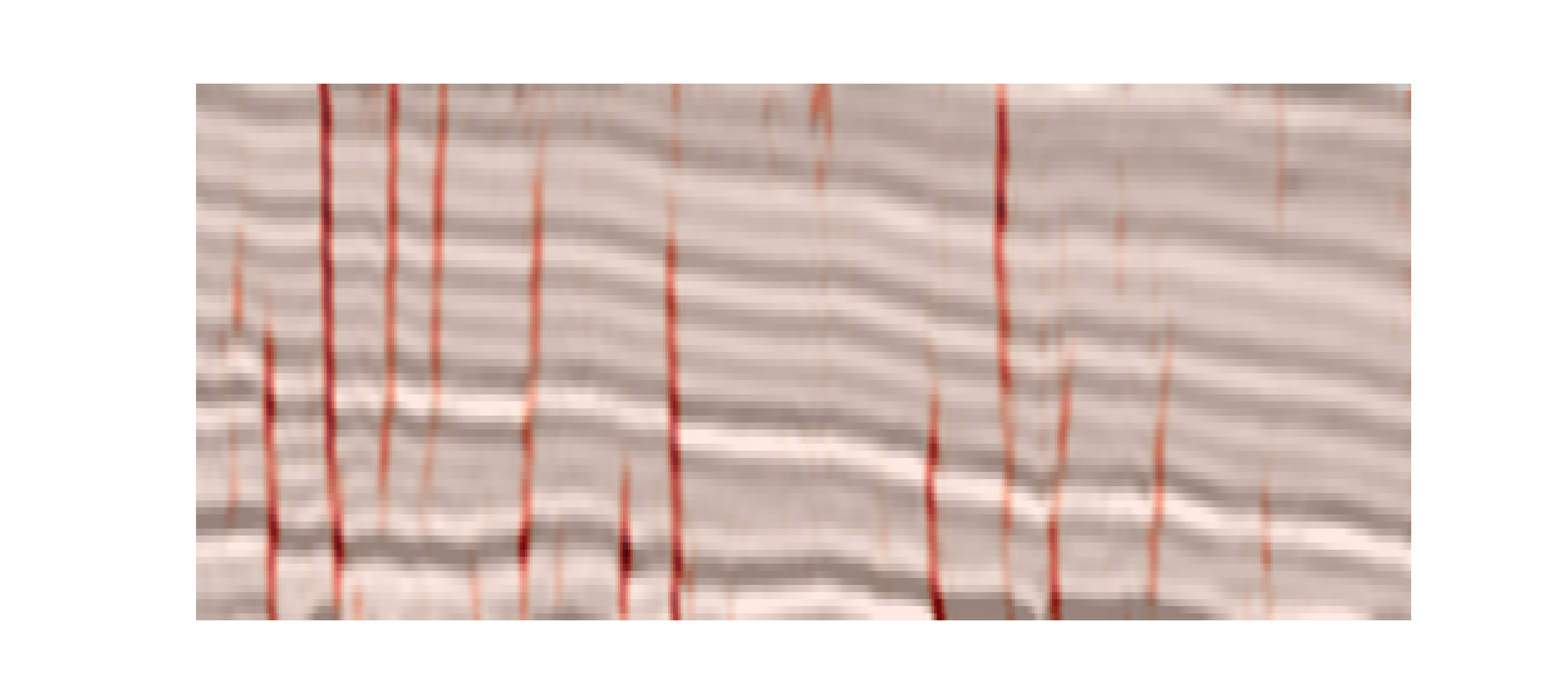}}
\label{fig_second_case_ts}
\caption{Comparison between our model and the previously annotated data for the crossline 2800 of GSB dataset. Note that the red regions represent the detections provided by our model, and despite diverging from interpretation of the expert in one of the annotated lines, it still demonstrates strong performance.
}
\label{fig_GSB}
\end{figure}

We also evaluated our model using the GSB dataset to ensure a fair comparison. The figure \ref{fig_GSB} illustrates the performance of our best model (self-attention) in detecting faults and heterogeneities for crossline 2800. The blue lines represent the manual interpretation provided by Haibin Di et al. \cite{di2017seismic}, while the image with the dark red lines depicts our results. Our model achieved an Intersection over Union (IoU) of 86\% and an F1 score of 88\% in this dataset.

\begin{figure*}[!t]
\centering
\includegraphics[width=3.8in]{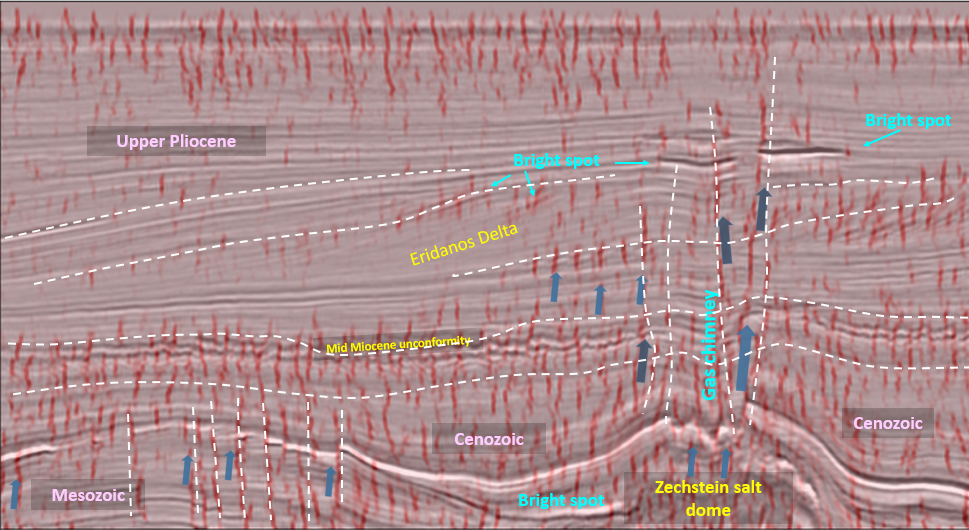}
\caption{Results performed over the F3 Block PSTM at inline 120. This image should be compared to Figure 1. Interesting observation suggestions on how different mechanisms of salt tectonics are related to the upward movement of gas, using structural paths to fill the traps, and generating bright spots. Blue arrows indicate these paths. White dotted lines suggest important rheological limits of layer-constrained faults and fractures.  }
\label{fig7}
\end{figure*}

We also made a qualitative assessment of our best results by applying the method to the entire F3 Dataset. The result obtained by our work allows the generation of different geological scenarios indicating that structural heterogeneities were isolated, which represent the primary fracture and fault systems individualized by regions in the study area. We have interpreted more structural details for Inline 120, as it is shown in Figure \ref{fig7} with outstanding results when our technique is applied to gas chimney regions, as previously illustrated in Figure 1. Results suggest how different mechanisms of salt tectonics are related to the upward movement of gas, using structural paths to fill the traps and generate bright spots. Note that the disturbances in the seismic signals that caused amplitudes with time pull-down under the gas chimney, do not affect the quality response of our algorithms. Attention is also asked to the algorithms’ response related to heterogeneities in the oldest Cenozoic reflection package, until the unconformity of the Mid-Miocene. Such unconformity looks like a rheological limit for layer-constrained faults and fractures. Also, in the Eridanos delta, significant heterogeneities are revealed showing chances of fluid retention and migration.

\begin{figure*}[!t]
\centering
\includegraphics[width=3.8in]{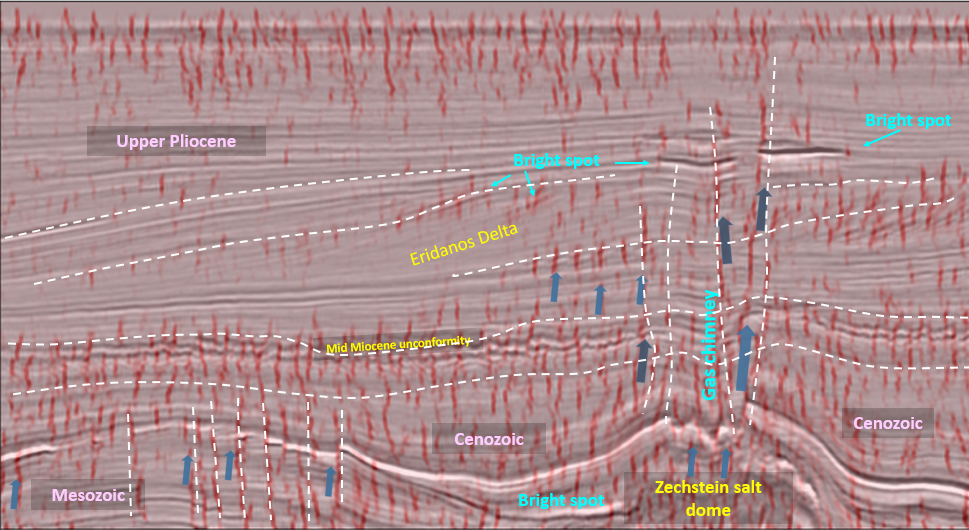}
\caption{Results performed over the F3 Block PSTM at inline 220. These images should be compared to those in Figure 3.  Amplitude anomalies related to gas leaks through extension faults suggest the geoscientific role of detected heterogeneities and gas migration paths. Blue arrows indicate these paths.  Dotted white lines suggest important rheological limits of layer-constrained faults and fractures}
\label{fig_sim220}
\end{figure*}

The second example is illustrated by Figure \ref{fig_sim220} where our workflow shows optimal performance to reveal amplitude anomalies related to gas leaks through extension faults applied to Inline 220. Detected heterogeneities show that a fault, despite could be modeled as a single object, becomes a numerical model and seismic mapping much easier, enhancing that, for many times, they should not be treated as a single structural event. It is also possible that an object described and modeled as a single fault, could be in fact the conjunction of several small faults, or restricted fracture zones, which may have fluid-sealing behaviors or, in other circumstances, have fluid-carrying behaviors within the same exploratory or production block. This is evident in the conical geometric pattern formed by the various heterogeneities that would make up modeled faults and can serve as ducts to feed trapped sediments. Once again, it is possible to observe the differential rheological patterns related to the tectonic deformation of pre-unconformity of Mid-Miocene sediments, and within the Eridanos delta. With the method here described, they are enhanced, suggesting that gas migration paths reach the seal unconformity at the top of the delta and then accumulated, also in small patches

\begin{figure*}[!h]
\centering
\includegraphics[width=3.8in]{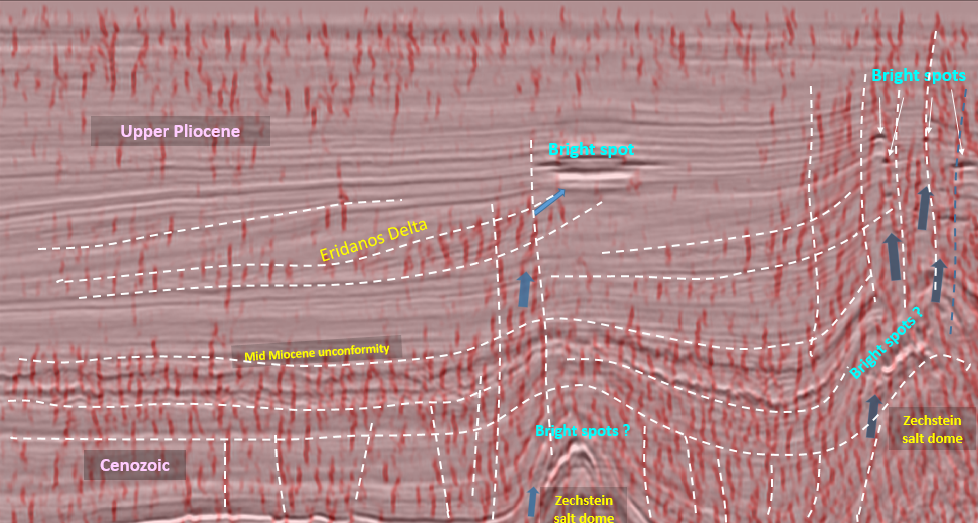}
\caption{Results performed over Inline 690 of the F3 Block in TWT (ms). We can see here how the structural heterogeneities are related to gas migration and its presence suggested by a bright spot over a flat spot. Blue arrows suggest possible gas paths. This is indicative of effective structural trapping for shallow gas in high porous stratigraphic intervals. These images should be compared to those in Figure 3.}
\label{fig_sim690}
\end{figure*}

The third example illustrates suggestions of different rheological regions, seen in Figure \ref{fig_sim690} where the white dashed lines show probable limits of differential interpretative elastic behavior for both stratigraphic zone and possible ducts and gas permeability barriers. The latter, more vertical, show how the gas may have loaded a trap highlighted by a shallow bright spot over a flat spot, and how its routes through heterogeneities could be extended up to the local reservoirs, pointed by the blue arrows in the figure. Note that all regions related to segments that could be modeled as faults, as suggested in Figure \ref{fig_4}, are densely represented, including many other heterogeneities with strong geological sense, which would not be hypothetically modeled, as those contained in the Erıdanos delta.

\section{Conclusion}
Earth structural heterogeneities have vital importance for the oil EP activities, once they affect the dynamics of subsurface fluid flow in hydrocarbon reservoirs, influencing operational, economic, and strategic decisions of the oil and gas industry. 
In this paper, we presented a novel and very effective approach to detect seismic structural heterogeneities using recent developments in Deep Learning architectures. We proposed changes to previous works by adding an Attention Block to an otherwise fully convolutional network. The advantage of using a SE-Net block is related to the ability to exploit channel-wise dependencies in the feature maps, at the cost of a slightly increased computational cost. The introduction of a Spatial Attention Layer also allowed the network to focus on local features to find any structures of interest. In contrast, the use of self-attention has a better performance where it attends jointly to both spatial and channel subspaces. Results show how our 2D solution is geoscientific important, computing efficient, and cost-effective to generate real economic and strategic impacts for E$\&$P Projects.
{
    \small
    \bibliographystyle{ieeenat_fullname}
    \bibliography{main}
}


\end{document}